\documentclass[a4paper, 10pt, conference]{ieeeconf}

\IEEEoverridecommandlockouts                              

\usepackage{xcolor}

\usepackage{graphics} 
\usepackage{epsfig} 
\usepackage{mathptmx} 
\usepackage{amsmath} 
\usepackage{amssymb}  
\usepackage{textgreek}
\usepackage{subcaption}
\usepackage{hyperref}
\usepackage{stfloats}
\usepackage{flushend}

\usepackage[belowskip=-10pt,aboveskip=12pt]{caption}
 
\title{\LARGE \bf Accelerated Sim-to-Real Deep Reinforcement Learning: Learning Collision Avoidance from Human Player}

\author{Hanlin Niu, Ze Ji, Farshad Arvin, Barry Lennox, Hujun Yin, and Joaquin Carrasco
\thanks{*This work was supported by EPSRC project No.EP/S03286X/1 and EPSRC RAIN project No. EP/R026084/1. \textit{(Corresponding author: Hanlin Niu)}}
\thanks{H. Niu,  F. Arvin, B. Lennox, H. Yin and J. Carrasco are with the Department of Electrical \& Electronic Engineering, The University of Manchester, Manchester, UK.
        {\{\tt\small hanlin.niu@manchester.ac.uk}\}}%
\thanks{Z. Ji is with the School of Engineering, Cardiff University, Cardiff, UK. 
        {\tt\small jiz1@cardiff.ac.uk}}

}

\begin{document}

\maketitle
\thispagestyle{empty}
\pagestyle{empty}

\begin{abstract}

This paper presents a sensor-level mapless collision avoidance algorithm for use in mobile robots that map raw sensor data to linear and angular velocities and navigate in an unknown environment without a map. An efficient training strategy is proposed to allow a robot to learn from both human experience data and self-exploratory data. A game format simulation framework is designed to allow the human player to tele-operate the mobile robot to a goal and human action is also scored using the reward function. Both human player data and self-playing data are sampled using prioritized experience replay algorithm. The proposed algorithm and training strategy have been evaluated in two different experimental configurations: \textit{Environment 1}, a simulated cluttered environment, and \textit{Environment 2}, a simulated corridor environment, to investigate the performance. 
It was demonstrated that the proposed method achieved the same level of reward using only 16\% of the training steps required by the standard Deep Deterministic Policy Gradient (DDPG) method in Environment 1 and 20\% of that in Environment 2. In the evaluation of 20 random missions, the proposed method achieved no collision in less than 2~h and 2.5~h of training time in the two Gazebo environments respectively. The method also generated smoother trajectories than DDPG. The proposed method has also been implemented on a real robot in the real-world environment for performance evaluation. We can confirm that the trained model with the simulation software can be directly applied into the real-world scenario without further fine-tuning, further demonstrating its higher robustness than DDPG. The video and code are available: \url{https://youtu.be/BmwxevgsdGc} \url{https://github.com/hanlinniu/turtlebot3_ddpg_collision_avoidance}
\end{abstract}

\section{Introduction}

Navigation is one of the key problems in autonomous mobile robots, involving a combination of algorithms for localisation, path planning and collision avoidance. To perform reactive behaviours in the presence of dynamic obstacles, the computational efficiency is an important criterion to evaluate navigation algorithms. 
Robot localisation can be achieved via various methods, including beacon-based methods, such as Wifi localisation, visible-light method and Ultra-wideband (UWB) solution, allowing navigation for robots without pre-calibrating the maps.
More popularly, autonomous navigation usually involves using Simultaneous Localisation and Mapping (SLAM) that allows a robot to build the map of an unknown environment beforehand or simultaneously using sensors like laser range finders or vision cameras. Path planning and collision avoidance can be then possible using a combination of global and local planners with the full or partial map information. However, it is time-consuming to build a complex and precise map using sensory data that is measured by tele-operating the robot to navigate around the environment. As well, autonomous navigation in highly dynamic and unstructured environment with moving obstacles remains challenging.

\begin{figure}
  \begin{center}
    \includegraphics[width=3.4in]{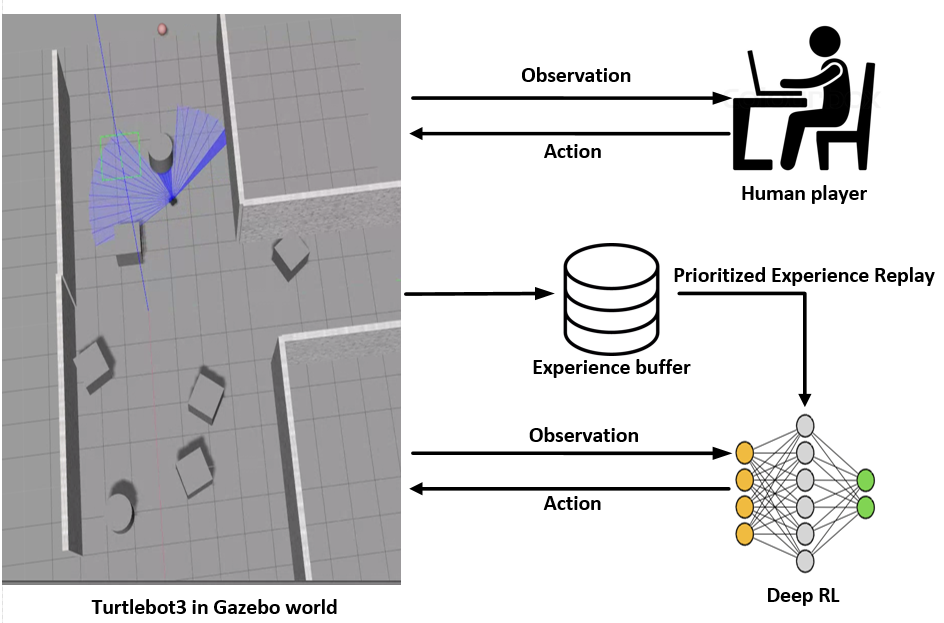}
    \caption{A learning-based mapless collision avoidance algorithm and training strategy were proposed by utilizing human demonstration data and simulated data.}
    \label{fig1}
  \end{center}
\end{figure}


The goal of this paper is to design a learning-based collision avoidance algorithm to generate continuous commands without the map and only use range data and the relative goal position to produce the reactive behaviour, as shown in Fig.~\ref{fig1}. Most deep reinforcement learning based algorithms are trained in simulation environment, including V-REP, PyGame , OpenAI, Gazebo, etc. The gap between the simulation environment and the highly complicated real-world environment is the main challenge to transfer the learned experience to the real mobile robot directly. In this paper, we use the laser data to abstract the distance between the mobile robot and the surroundings. We also configure the simulation in a game-like environment, allowing human players to tele-operate the mobile robot to avoid the obstacles and navigate to the goal. A reward function is designed to score the human action. A human player data logger is designed to record the human action data, robot state data and reward data. The human experience and robot experience are then mixed. Prioritised experience replay approach \cite{vecerik2017leveraging} is applied to make the robot to learn human data and robot data more efficiently.

The fast learning capabilities of the proposed approach is validated using a turtlebot 3 mobile robot equipped with a laser sensor in both simulation and real-world environments. The method is compared with the standard DDPG policy, which only learns from the simulated data. The main contribution of this paper is three-fold: 1) A deep reinforcement learning-based mapless collision avoidance algorithm is proposed; 2) an efficient learning strategy is proposed by enabling the autonomous agent to learn from both simulated agent and human player and the number of training step is only 16\% and 20\% of standard DDPG in two simulation environments respectively; and 3) the trained policy from the simulated environment can be deployed onto the real mobile robot without further fine-tuning. 4) The code of the proposed algorithm and framework for training robot is released as open-source.

\section{Related Work}

With the development of computational hardware and access to large amount of training data, deep learning shows a great potential in solving computer vision and navigation problems. A decentralized collision avoidance policy is proposed in \cite{long2018towards}, where a multi-scenario multi-stage training framework is adopted to train the policy. It is demonstrated that the learned policy can be used for navigating a large number of robots in the simulated environment. A successor-feature-based deep reinforcement learning algorithm is proposed in \cite{zhang2017deep} that can reduce the training time, when transfer knowledge from the previously learned experience to the new scenarios. Tai \textit{et al.} \cite{tai2017virtual} proposed an end-to-end learning-based mapless motion planner by using laser data and the relative target information as input data. The policy outputs command data directly and its efficiency is demonstrated in unknown simulated and real scenarios. A novel deep reinforcement learning method, integrating an LSTM agent and Local-Map Critic, is proposed by Choi \textit{et al}. \cite{choi2019deep}. This method enables the mobile robot to navigate in a complex environment with limited field of view. The mobile robot, using NVIDIA Jetson TX2 as a processor and Intel Realsense D435 with a FOV of $90^\circ$, outperforms the methods having a wider FOV. A value based network is applied to generate velocity commands in real-time and also considers the uncertainty of other agents. Comparing with optimal reciprocal collision avoidance, the proposed method shows more than 26\% improvement in terms of the required time to reach the goal. Deep reinforcement learning is not only applied to avoid collision in near field, a long-term path planning algorithm is proposed by Zhu \textit{et al.} \cite{zhu2018deep} to navigate the agent to visit unexplored environments. It is realized by integrating the deep reinforcement learning model with a greedy strategy.

Deep reinforcement learning has also been implemented on social mobile robots. A social norms-based deep reinforcement learning algorithm is proposed by Chen \textit{et al.} \cite{chen2017socially} that models subtle human behaviors and navigation rules for robot navigation, including passing, crossing and overtaking. The mobile robot is demonstrated to navigate through a pedestrian-rich environment at a human walking speed. Everett \textit{et al.} \cite{everett2018motion} proposes a strategy using LSTM to enable the autonomous agent to navigate through an arbitrary number of agents, extending the previous work that can only work with fixed number of agents. The attention-based motion planner and deep reinforcement learning algorithm are integrated by Chen \textit{et al.} \cite{chen2019crowd}, which enables the agent to learn from the Human-Human interactions in crowd. It is demonstrated that the agent can finally anticipate human behaviours and navigate itself through the crowds. A composite reinforcement learning framework is proposed by Ciou \textit{et al.} \cite{ciou2018composite}. The proposed algorithm makes the system possible to keep collecting human feedback to adjust the rewards to the social norms, which enables the agent to modify its velocity and perform Human-robot interaction in social environment. Kohari \textit{et al.} \cite{kohari2018generating} applied deep deterministic policy gradient algorithm (DDPG) on attendant robot to generate adaptive behaviors based on the classification states of human, including walking, standing, sitting and talking. 

\raggedbottom

Deep reinforcement learning algorithm has demonstrated its superior in reducing computational time to traditional control algorithms, such as MPC (Model Predictive Control) \cite{zhang2016learning}, validated on unmanned aerial vehicles for fast maneuvers. A fast reactive navigation algorithm is proposed by Sampedro \textit{et al.} \cite{sampedro2018laser} and the algorithm applies 2D-laser range measurements and the relative goal position as the input for the deep neural network. A DJi Matrice 100 quadcopter, equipped with Hokuyo laser rangefinder UTM-30LX, is trained in a Gazebo-based 3D simuation environment. The learned experience is carried out in an unknown indoor environment, with the presence of static and dynamic obstacles. The DDPG algorithm is applied by Rodriguez-Ramos \textit{et al.} \cite{rodriguez2018deep} to enable a Parrot Bebop 2 drone to land on a moving platform. The drone model is trained in the Gazebo 3D environment and Rotors UAV simulator and then applied in the real-world environment. Haksar \textit{et al.} \cite{haksar2018distributed} proposed a computationally efficient algorithm using Multi-Agent Deep Q Network (MADQN) to control a swarm of UAVs to fight forest fires. Its scalability is demonstrated by carrying out missions with various forest sizes and different number of UAVs with various model parameters. 

\raggedbottom

The algorithms mentioned above are mainly trained using the data from a simulated agent. However, learning from a non-experienced agent makes the learning process inefficient. This research enables the proposed mapless collision avoidance algorithm not only learn from simulated agent but also from human player, making the learning process faster and better.

\section{Problem Formulation}

The collision avoidance problem for mobile robots is defined in the context of an autonomous agent moving on the Euclidean plane in the presence of obstacles. At each time step $t$, the robot has access to the surrounding observation information state $s_{t}$ and generates the action velocity command $v_{t}$ that allows collision free navigation for the robot to the goal. The relative position to the goal is denoted by $g_{t}$. Instead of perfect observation of the whole map, we assume that our robot has only partial observation in the range of the laser sensor and the relative goal position. This assumption makes our method more practical and robust in the real world environment. The velocity $v_{t-1}$ of last time step is also considered. Therefore, the problem can be formulated as to find the translation function:

\begin{equation} \label{eq1}
\ v_{t} = f\left( s_{t}, g_{t}, v_{t-1} \right)
\end{equation}

\section{Approach}
\subsection{Reinforcement learning Setup}

\subsubsection{Observation space and action space}
The observation state $s_{t}$ consists of laser data $l_{t}$, velocity $v_{t}$ and relative target position $g_{t}$. $l_{t}$ is a 1D array laser information and it is normalized by its maximum range. The velocity data $v_{t}$ represents the linear velocity and the rotational velocity of the agent. The relative target position $g_{t}$ is calculated in the polar coordinate (relative distance and angle to the current position). The neural network outputs continuous velocity commands to the robot. The action space of neural networks consists of linear velocity $l_v$ and the angular velocity $a_v$.

\subsubsection{Reward space}
As the mobile robot should navigate to the goal, while keeping clear distance from obstacles, the reward is set with the consideration of the clearance distance and the step distance towards to the goal. The reward function is given as:

\begin{equation}
\ r = r_g + r_c + r_{av} + r_{lv} \
\label{eq2}
\end{equation}

where $r$ stands for the total reward, $r_g$ denotes the distance reward, $r_c$ represents the collision reward, $r_{av}$ and $r_{lv}$ represent the angular velocity punishment (negative reward) and linear velocity punishment, correspondingly. $r_g$ is calculated using:

\begin{equation}
r_g = \begin{cases} 
r_{arrival}  & \quad \text{if} \,\, d_g < d_{gmin}\\
\Delta {d_g}  & \quad otherwise
\end{cases}
\label{eq3}
\end{equation}

where $d_g$ denotes the distance to the goal. When $d_g$ is less than a threshold $d_{gmin}$, the robot is considered to arrive the goal and it will receive a reward $r_{arrival}$, otherwise it will receive reward that equals to the distance it travelled during the last time step. The collision reward $r_c$ can be calculated by: 

\begin{equation}
r_c = \begin{cases} 
r_{collision} & \quad \text{if} \,\, d_{romin} \leq d_{ro} < 2d_{romin} \\
2r_{collision} & \quad \text{if} \,\, d_{ro} < d_{romin}   \\
0  & \quad otherwise
\end{cases}
\label{eq4}
\end{equation}

where $d_{ro}$ denotes the distance between robot and obstacles and $d_{romin}$ represents the threshold value of $d_{ro}$. $r_{collision}$ denotes the collision reward. Negative reward $r_{av}$ and $r_{lv}$ are introduced, which will ensure the algorithm producing smooth trajectories, and they can be calculated by:

\begin{equation}
r_{av} = \begin{cases} 
r_{ap} & \quad \text{if} \,\, \left|a_{v}\right| > 0.8\left|a_{vmax}\right| \\
0  & \quad otherwise
\end{cases}
\label{eq5}
\end{equation}

\begin{equation}
r_{lv} = \begin{cases} 
r_{lp} & \quad \text{if} \,\, l_{v} < l_{vmin} \\
0  & \quad otherwise
\end{cases}
\label{eq6}
\end{equation}

where $a_{vmax}$ denotes the maximum angular velocity threshold and $l_{vmin}$ represents the minimum linear velocity threshold. The robot will receive punishment reward $r_{ap}$ or $r_{lp}$, when its angular velocity $\left|a_v\right|$ is larger than threshold $0.8\left|a_{vmax}\right|$ or when linear velocity $l_v$ is lower than threshold $l_{vmin}$. Note that although sigmoid function, which has an asymptotic limit, is used for generating angular velocity command, we still add a second threshold to limit the angular velocity in a smaller range, which makes it possible to maneuver in a fast angular velocity although it is not encouraged.

\subsection{Network architecture}
\begin{figure}[!ht]
  \begin{center}
    \includegraphics[width=3.4in]{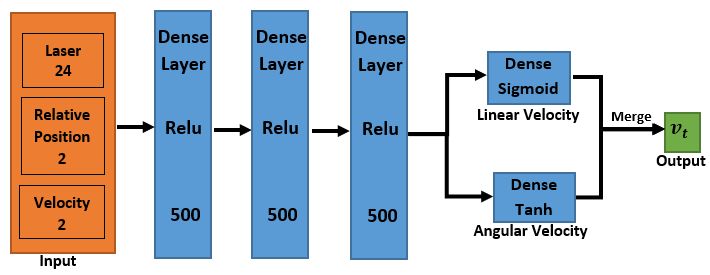}
    \caption{The architecture of the actor network}
    \label{fig2}
  \end{center}
\end{figure}

In this work, the DDPG algorithm is applied as the base of the proposed method. As shown in Fig.~\ref{fig2}, the input of the actor network is a 28-dimensional input vector, comprising the 24-dimensional laser data, merged with the 2-dimensional relative goal information and 2-dimensional robot action information. The input data is connected with three dense layers (500 nodes each) and the actor network produces the linear velocity and angular velocity by using a Sigmoid function and a hyperbolic tangent function respectively. These two velocity commands are merged as the action input of the critic network. The state input of the critic network is the same as the input of the actor network and it is processed by three dense layers (500 nodes each). The action input is integrated with the second dense layer. Finally, the critic network generates the Q-value through a linear activation function.

\begin{figure}[!ht]
  \begin{center}
    \includegraphics[width=3in]{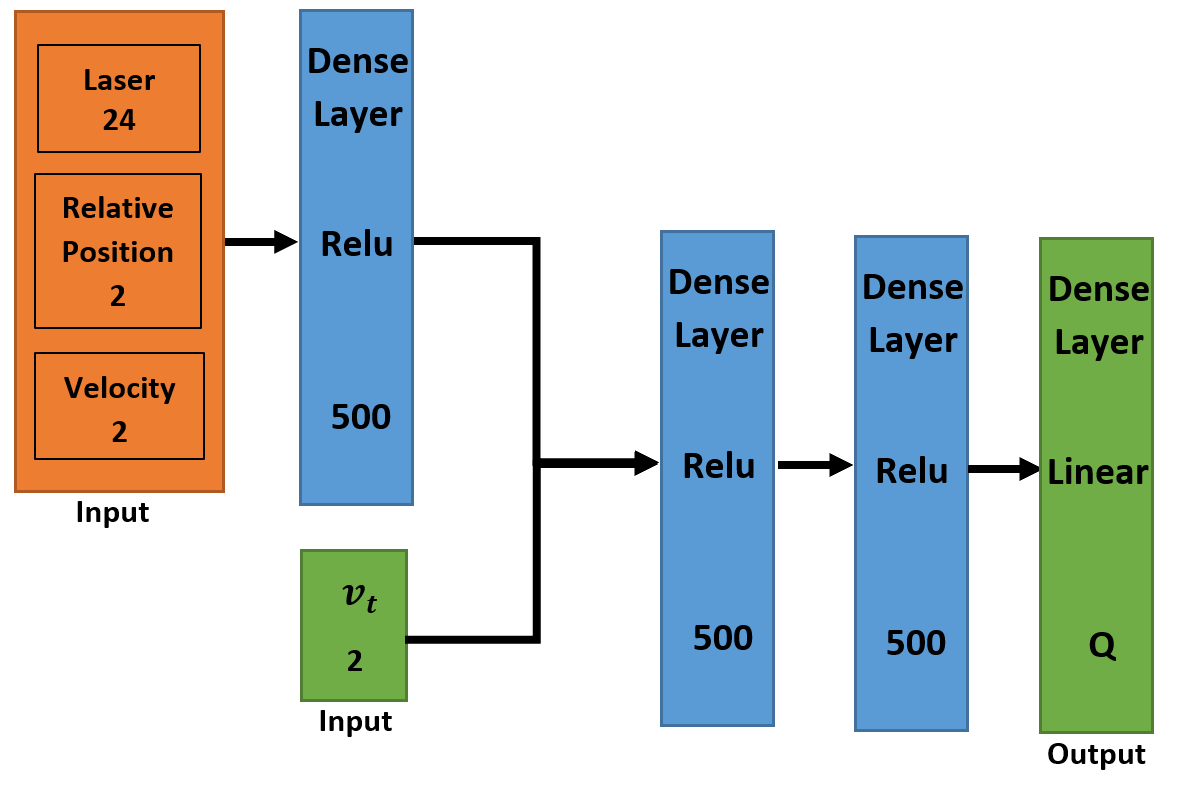}
    \caption{The architecture of the critic network}
    \label{fig3}
  \end{center}
\end{figure}

\subsection{Human player data collection}
\begin{figure}[tb]
  \begin{center}
    \includegraphics[width=3.0in]{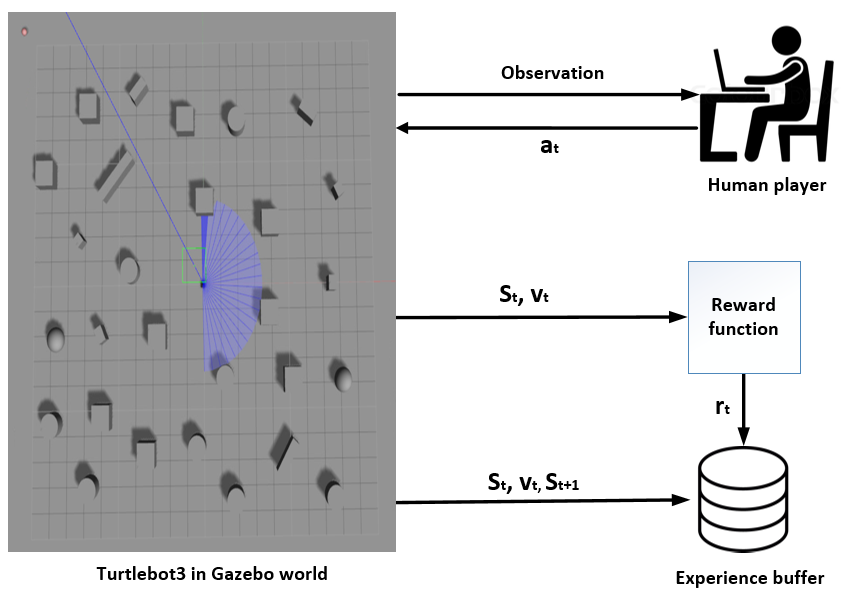}
    \caption{Human player data collection}
    \label{fig4}
  \end{center}
\end{figure}

The demonstration data is collected through tele-operating the robot by human players, as shown in Fig.~\ref{fig4}. The laser data $l_{t}$, velocity data $v_{t}$ and relative target position data $g_{t}$ are merged as the current state $s_{t}$ and stored. $v_{t}$ is performed by Human. The next time step state $s_{t+1}$ is recorded by reading the robot state from gazebo environment directly. $s_{t}$,  $v_{t}$ and $s_{t+1}$ are used to evaluate the human player's score $r_{t}$, using the same reward function designed for the auto mission. Finally, $s_{t}$, $v_{t}$, $s_{t+1}$ and $r_{t}$ are stored in the experience replay buffer, allowing the agent to maintain these transitions to propagate the rewards throughout the value function.

\subsection{Prioritized experience replay}

Instead of sampling historical data uniformly, the Prioritized Experience Replay algorithm (PER) \cite{vecerik2017leveraging} is applied to ensure important historical data to be sampled more frequently. Along with the experience buffer, priority information of each transition is calculated by \eqref{eq7} and is stored in the priority tree. 

\begin{equation}
\ P_i = \frac{p^{\alpha}_{i}}{\sum\nolimits_{k} p^{\alpha}_{k}} \
\label{eq7}
\end{equation}

where the sampling probability of the i$^{th}$ transition is denoted by $P_i$, $\alpha$ denotes distribution factor and $p_i$ stands for the priority of a set of transition data, formulated in \eqref{eq8}.

\begin{equation}
\ p_i = {\delta}^{2}_{i} + \lambda \left|\triangledown_a{Q}(s_i,a_i|\theta^Q)\right|^2 + \epsilon + \epsilon_{D} \
\label{eq8}
\end{equation}

where $\delta$ stands for the TD (time difference) error, the second term $\lambda \left|\triangledown_a{Q}(s_i,a_i|\theta^Q)\right|^2$ denotes the actor loss, $\lambda$ is a contribution factor and $\epsilon$ represents a small positive sampling probability for each transition, which ensures each transition data can still be sampled. $\epsilon_D$ is applied to increase the sampling probability of the demonstration data. Each transition is evaluated by :

\begin{equation}
\ \omega_i =(\frac{1}{N} \cdot \frac{1}{P_i})^\beta \
\label{eq9}
\end{equation}

where $\omega_i$ denotes the sampling weight for updating the network, indicating the importance of each transition data, $N$ stands for the batch size, and $\beta$ is a constant value for adjusting the sampling weight.

\section{Experiments and results}

\subsection{Training and experiment setup}
\begin{figure}[tb]
  \begin{center}
    \includegraphics[width=2.6in, height=1.8in]{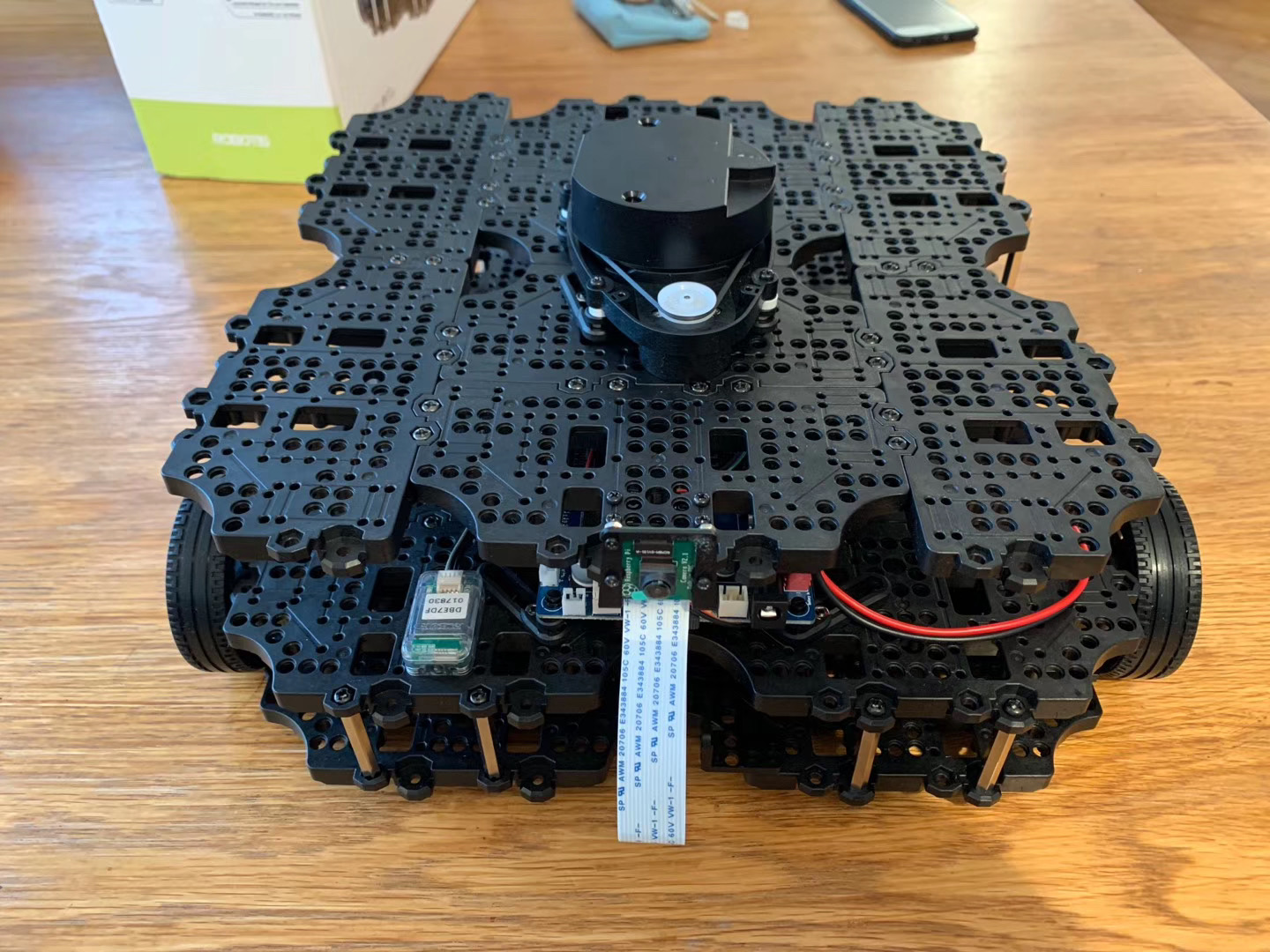}
    \caption{TurtleBot3 Waffle Pi robotic platform.}
    \label{fig5}
  \end{center}
\end{figure}

\begin{figure*}
     \centering
     \begin{subfigure}[b]{0.32\textwidth}
        \centering
        \includegraphics[width=2.0in, height=1.7in]{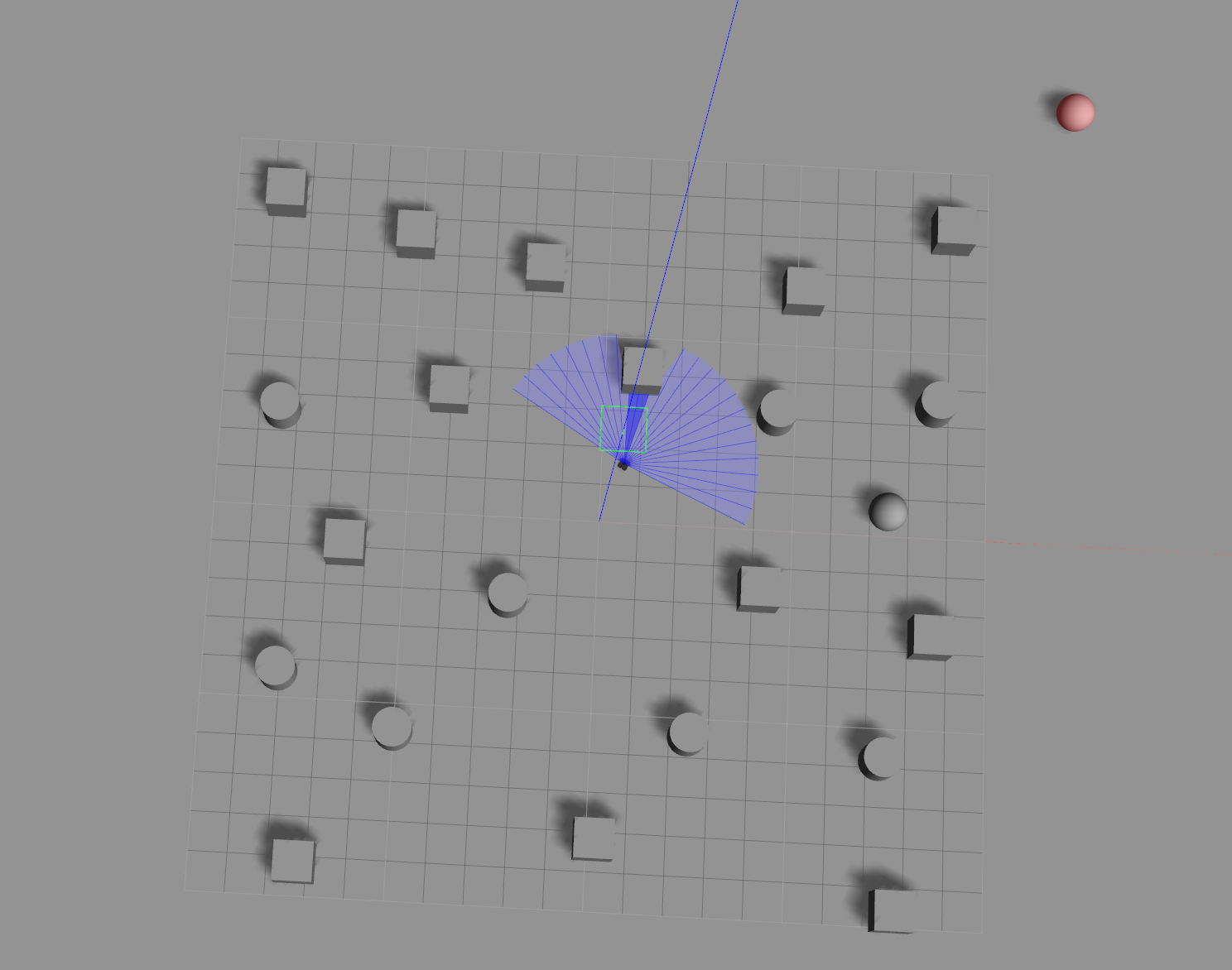}
        \caption{ }
        \label{fig6}
     \end{subfigure}
     \hfill
     \begin{subfigure}[b]{0.32\textwidth}
         \centering
         \includegraphics[width=2.4in, height=1.8in]{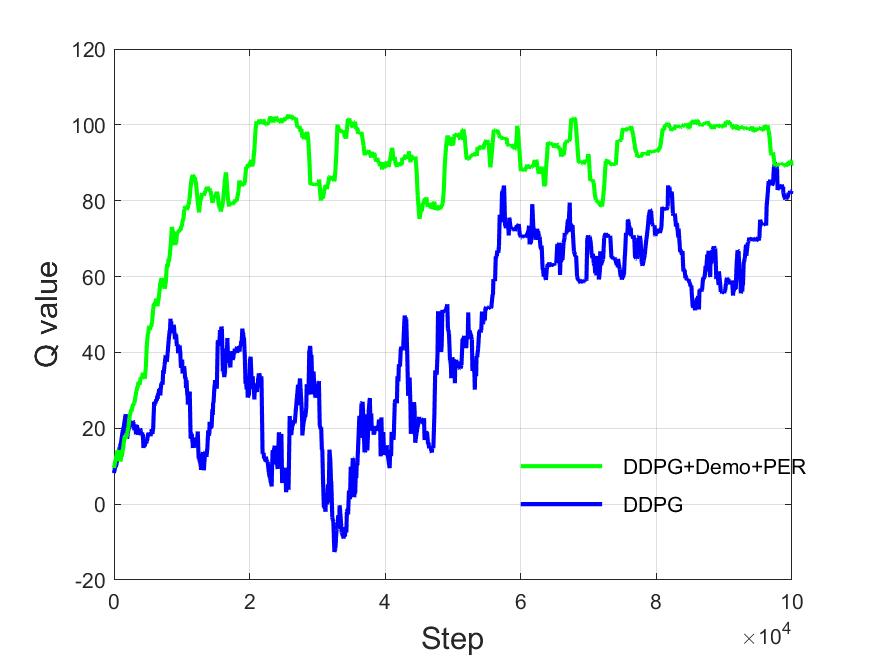}
         \caption{ }
         \label{fig8}
     \end{subfigure}
     \hfill
     \begin{subfigure}[b]{0.32\textwidth}
         \centering
         \includegraphics[width=2.4in, height=1.8in]{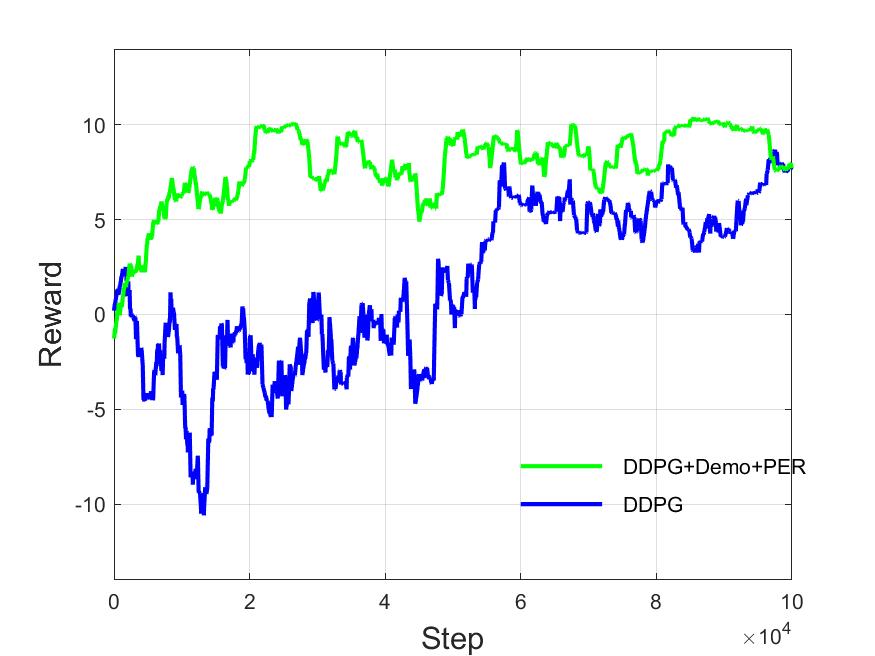}
         \caption{ }
         \label{fig9}
     \end{subfigure}
        \caption{(a) Environment 1 in Gazebo, (b) Q-value comparison and (c) reward comparison of the proposed method with DDPG in Environment~1.}
        \label{fig6sum}
\end{figure*}

\begin{figure*}
     \centering
     \begin{subfigure}[b]{0.32\textwidth}
        \centering
        \includegraphics[width=2.0in, height=1.7in]{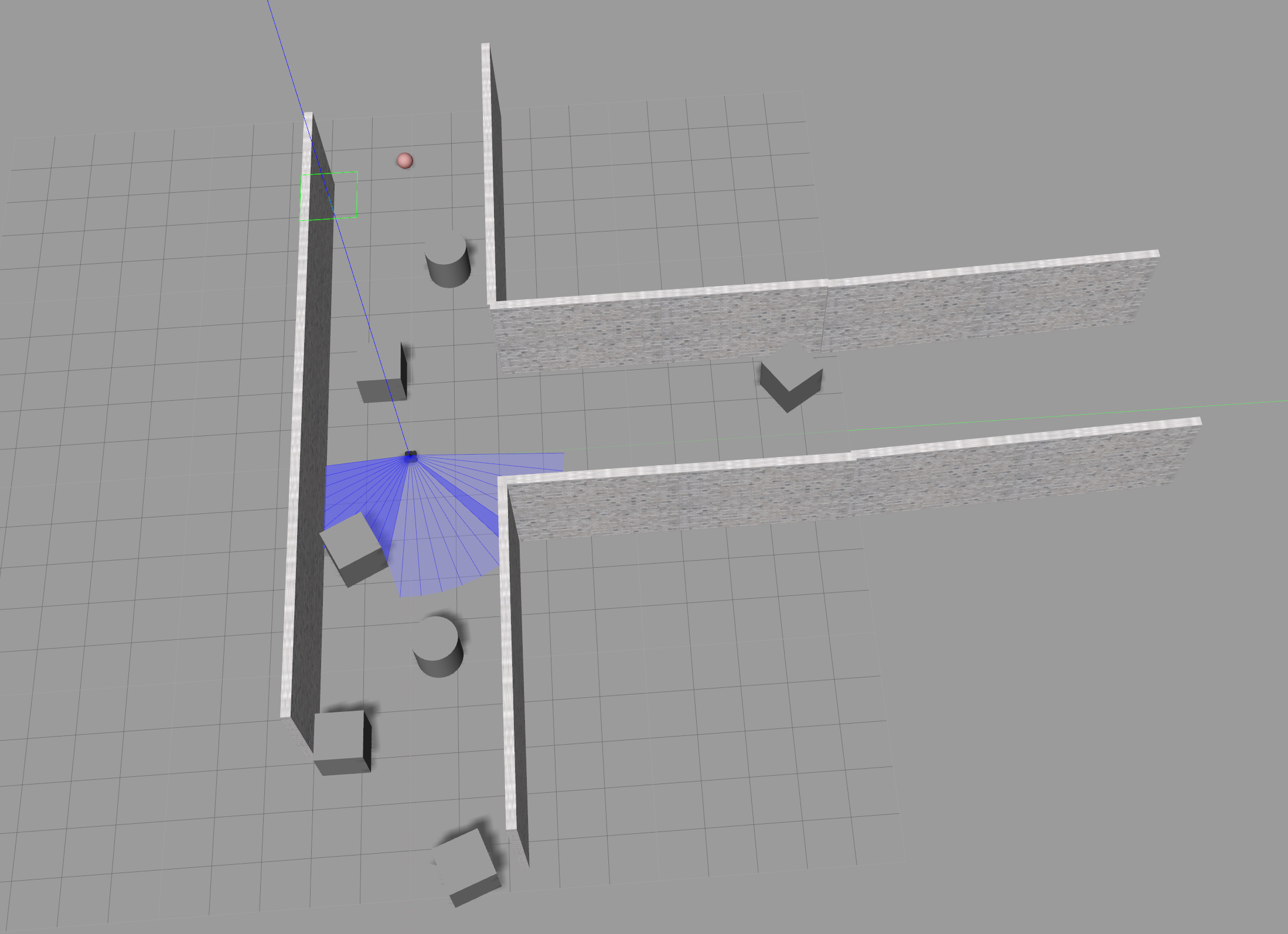}
        \caption{ }
        \label{fig7}
     \end{subfigure}
     \hfill
     \begin{subfigure}[b]{0.32\textwidth}
         \centering
         \includegraphics[width=2.4in, height=1.8in]{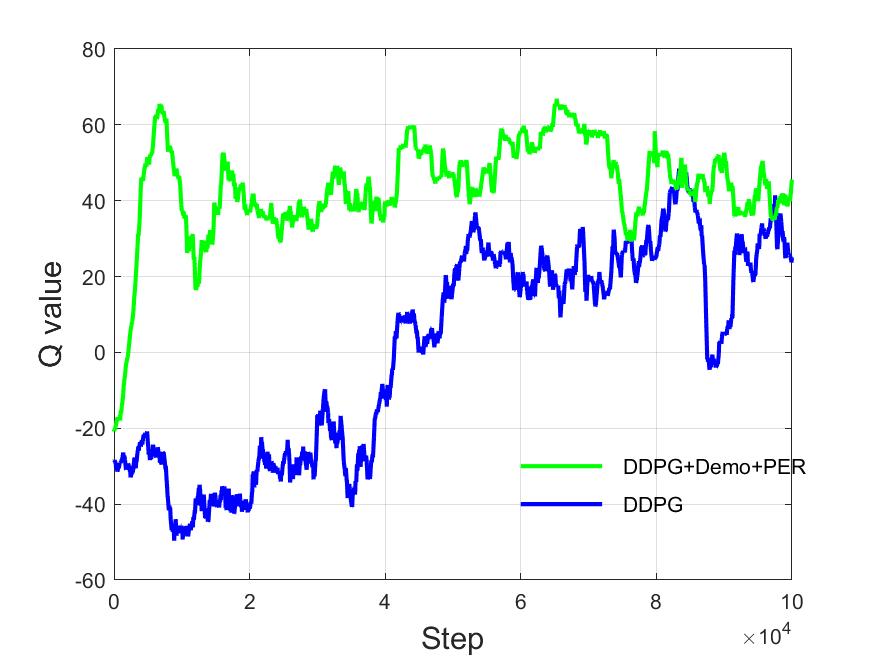}
         \caption{ }
         \label{fig10}
     \end{subfigure}
     \hfill
     \begin{subfigure}[b]{0.32\textwidth}
         \centering
         \includegraphics[width=2.4in, height=1.8in]{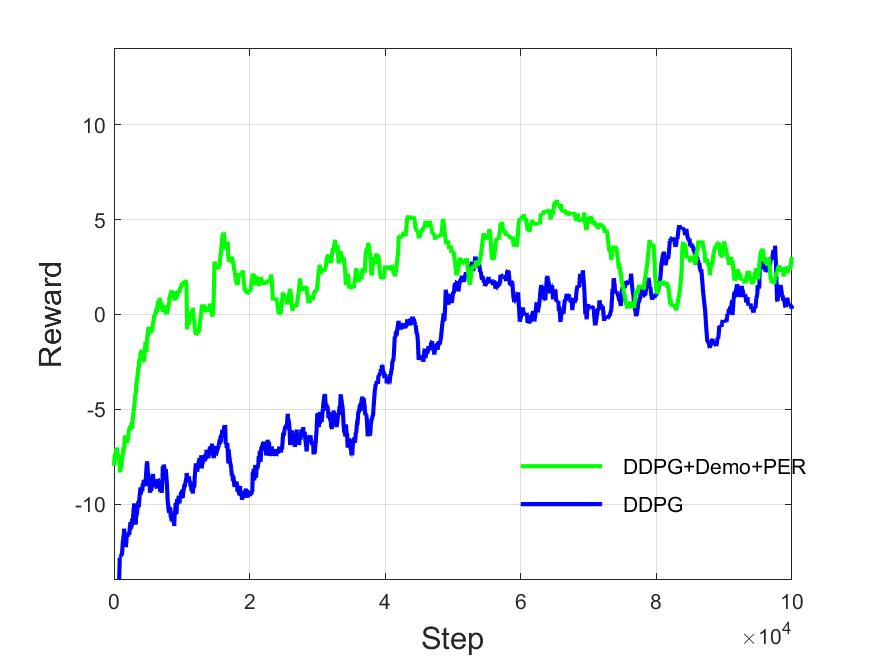}
         \caption{ }
         \label{fig11}
     \end{subfigure}
        \caption{(a) Environment 2 in Gazebo, (b) Q-value comparison and (c) reward comparison of the proposed method with DDPG in Environment~2.}
        \label{fig7sum}
\end{figure*}

The TurtleBot3 Waffle Pi mobile robot was used in both simulation and real-world environment, as shown in Fig.~\ref{fig5}. The training process was undertaken in the Gazebo environment, which minimises the effort to transfer the learned experience from simulation to the real-world real-robot experiment, as both simulated and real models use the same ROS interface and Gazebo world also simulates the dynamics of the robot. The robot model was trained in a clustered environment (Environment 1) and a narrow corridor environment (Environment 2), as shown in Fig.~\ref{fig6} and Fig.~\ref{fig7}. The small red ball was used to represent the target and was randomly placed in each episode. The blue lines visualize the laser information around the robot. The robot subscribes to the laser data from a 360 Laser Distance Sensor LDS-01, which has a $360^\circ$ field of view and its range is $(0.12~m, 3.5~m)$. In this work, we only use 24 laser readings from ($-90^\circ$, $90^\circ$), as we assume the robot will only move forward.

The training was carried out on a computational platform with a NVIDIA TITAN RTX GPU. The model was trained using Adam optimiser \cite{kingma2014adam}. The learning rate of actor and critic network were set to 0.0001.

\subsection{Training evaluation}
The standard DDPG algorithm was implemented for comparison. The standard DDPG algorithm used the same actor and critic network as the proposed method. The difference is the standard DDPG method learns from scratch and samples uniformly from the history data without taking into consideration of priority. The proposed method not only uses human demonstration data but also implements the PER algorithm. In this research, 1000 sets of transition data from human demonstrators were recorded for both Environment 1 and Environment 2 respectively, which cost around 20 minutes.

\subsubsection{Evaluation in simulated environment}
Q-value and reward of both the DDPG and the proposed method in the training procedure were recorded for comparison, as shown in Fig.~\ref{fig6sum} and Fig.~\ref{fig7sum}. The mean value of Q-value and reward of 4000 continuous steps are applied to reduce the effect of the random mission problem. In Environment 1, the Q-value and reward of the proposed method increases much faster than the standard DDPG. The Q-value of the proposed method reaches and maintains its level in the range from 80 to 100 in less than 10000 steps, while the standard DDPG achieves 80 after around 60000 steps and its Q-value stays between 60 and 80 afterwards. The trend of the reward is similar with Q-value. The reward of the proposed method stays between 5 and 10 after only about 10000 steps and the standard DDPG requires 60000 steps to reach the range of only from 4 to 8. Fig.~\ref{fig8} and Fig.~\ref{fig9} indicate that the proposed method requires only $16\%$ of the training step number with the standard DDPG and achieves better performance. In Environment 2, the proposed method still achieves higher Q-value and reward faster than DDPG. It is found that the Q-value and reward are lower than those in Environment 1, because the robot has to navigate through narrow obstacles and ends with lower collision reward. As shown in Fig.~\ref{fig10}, the Q-values of both the proposed method and DDPG reaches 20 with about 10000 steps and, in contrast, 50000 steps respectively. The average Q-value of the afterwards steps of the proposed method is also higher than DDPG. The reward has the same trend as Q-value, as shown in Fig.~\ref{fig11}. It can be concluded that the proposed method only uses 20\% of the training step number required by the standard DDPG and achieves higher Q-value and reward in Environment 2.




Twenty random missions are used to evaluate the performance of two algorithms. In Environment 1, it is found the proposed method achieves no collision in 20 missions only after 25000 steps, which costs only 1 hour and 53 minutes for training. The standard DDPG still encountered 7 collisions after 60000 training steps and 1 collision after 70000 training steps. It is also noted that the proposed method generated smoother trajectories and navigated the robot to the goal in a shorter time. In Environment 2, the proposed method achieves no collision in 20 missions after 35000 steps, which cost 2 hours and 25 minutes for training, while the standard DDPG still had 3 collisions in 20 trials after 80000 steps and it also had unnecessary turning manoeuvres. 

\raggedbottom

\subsubsection{Evaluation in real-world environment}
\begin{figure*}[t]
     \centering
     \begin{subfigure}[b]{0.3\textwidth}
         \centering
         \includegraphics[width=2.1in, height=1.6in]{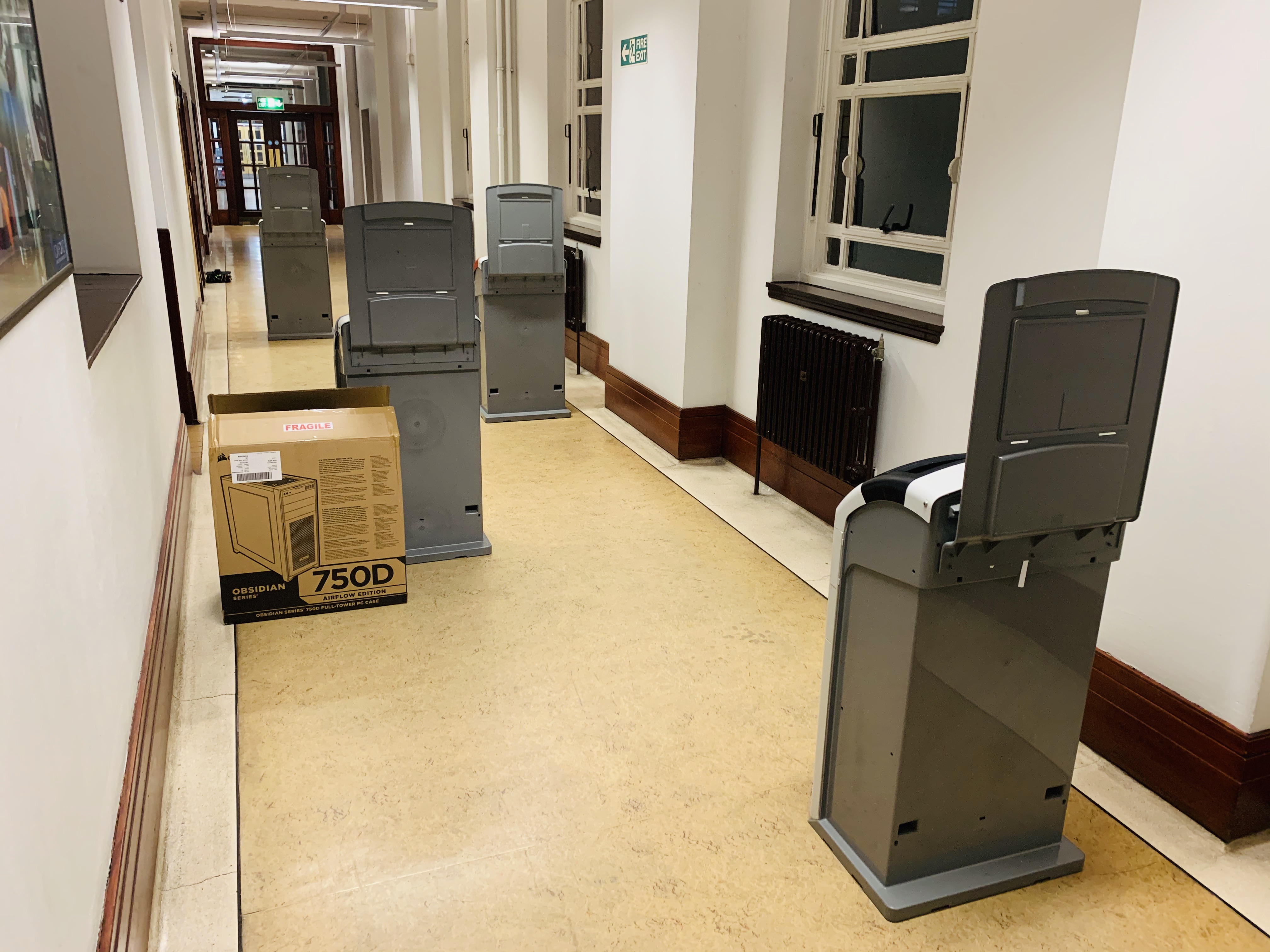}
         \caption{Corridor of Sackville Street Building}
         \label{fig12}
     \end{subfigure}
     \hfill
     \begin{subfigure}[b]{0.3\textwidth}
         \centering
         \includegraphics[width=2.1in, height=1.6in]{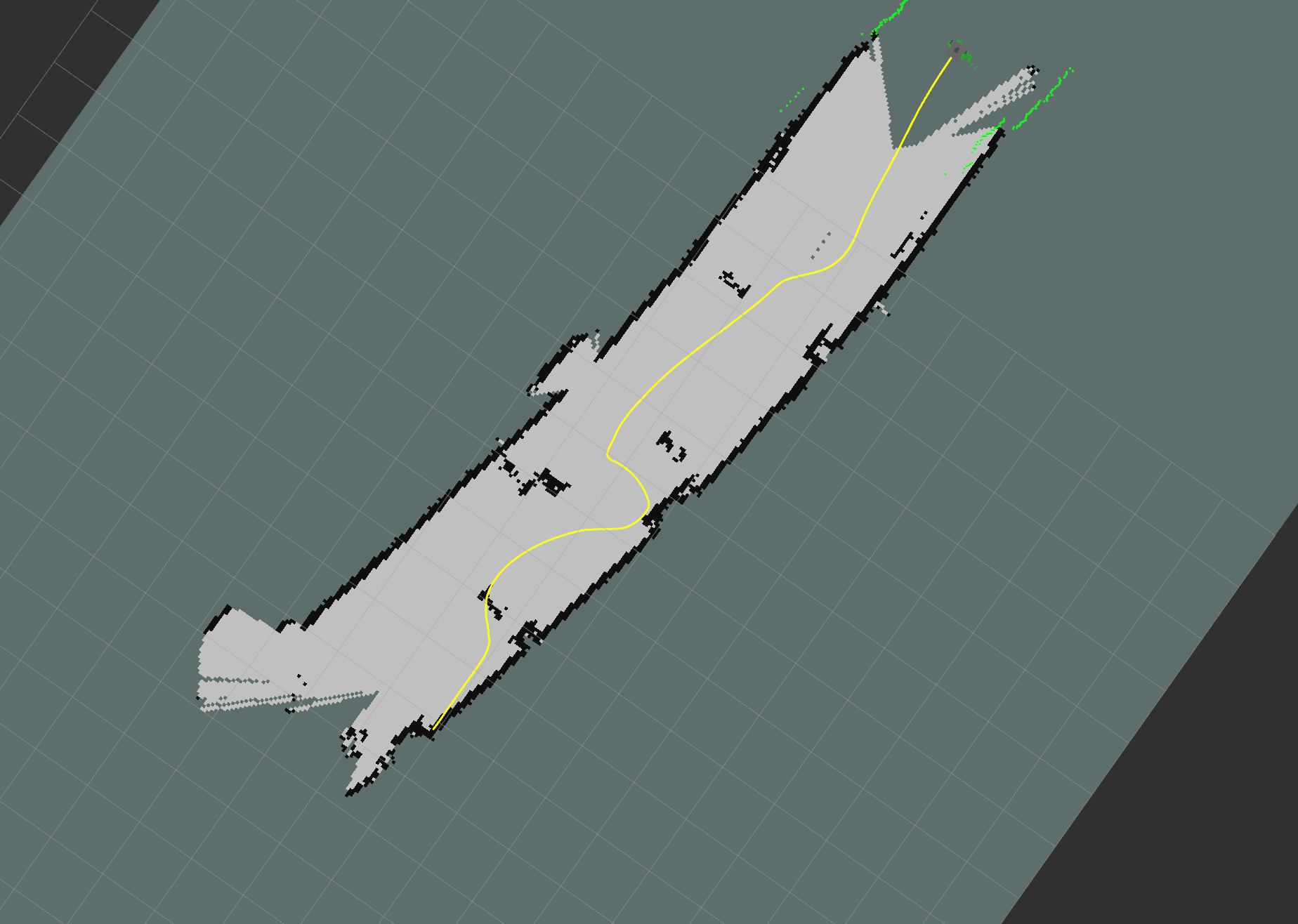}
         \caption{Trajectory of the proposed method}
         \label{fig13}
     \end{subfigure}
     \hfill
     \begin{subfigure}[b]{0.3\textwidth}
         \centering
         \includegraphics[width=2.1in, height=1.6in]{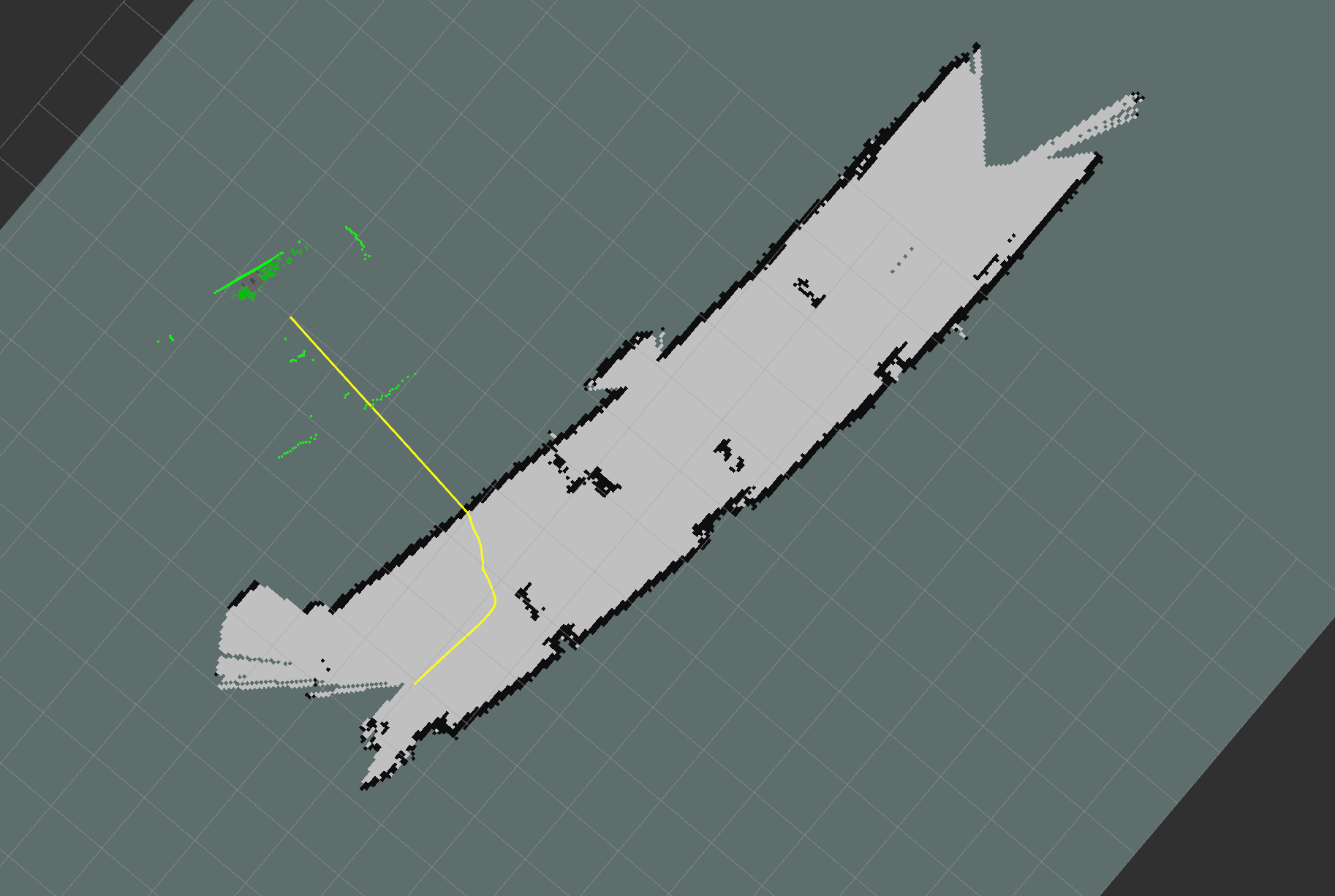}
         \caption{Trajectory of standard DDPG}
         \label{fig14}
     \end{subfigure}
        \caption{Comparison of the proposed method and DDPG in real-world environment}
        \label{fig15}
\end{figure*}

To test the performance of transferring the learned knowledge directly from simulation to the real world, the models trained from the Gazebo Environment 2 were tested in the corridor of Sackville Street Building in the University of Manchester, as shown in Fig.~\ref{fig12}. Four bins were placed in the middle of the corridor to test the collision avoidance capability of both algorithms. The final goal was set behind of the fourth bin. In Fig.~\ref{fig13} and Fig.~\ref{fig14}, the yellow lines represent the trajectories of the robot. The proposed method navigated the robot to pass all the four obstacles, while using the standard DDPG, the robot collided with the wall at the first turn. Five sets of the same task were repeated and the proposed method completed all missions and none of the mission can be achieved by DDPG. Note that, the robot position and the goal position were estimated using the TurtleBot3's built-in AMCL algorithm. The map was only used for demonstration. The robot only senses the obstacles in its laser detection range and the decision is made by its current sensor reading. In Fig.~\ref{fig13}, the yellow trajectory crossed the first bin, which indicates our robot could achieve the collision avoidance capability in presence of the inaccuracy of the localisation algorithm. We also tested to put sudden obstacle in front of robot to test its obstacle avoidance capability in dealing with dynamic environment, the robot could still avoid the obstacle. 

\raggedbottom

\section{Conclusion and Future Work}

In this paper, a deep reinforcement learning based collision avoidance algorithm was proposed. The proposed method uses human demonstration data and the prioritized experience replay algorithm to improve the performance of the standard DDPG algorithm. In simulated Environment 1, the training step number of the proposed method was only $16\%$ of the required by the DDPG algorithm and achieved higher reward and Q-value overall. In Environment 2, the proposed method only needed $20\%$ training step number of standard DDPG. A quantitative comparison was also given by comparing the two algorithms in 20 random missions. It was found that the proposed method needs less than 2~h training time to achieve no collision in Environment 1 and 2.5~h in Environment 2, while still generates smoother trajectory than standard DDPG. The proposed method was also tested in the unknown real-world environment without further fine-tuning. It was found that the proposed method was more robust than DDPG in real-world applications. In the future work, RGB images and depth information will be considered to analyse and predict the behaviours of pedestrian hence the behaviour of the robot can be adjusted with human preference. Also, the LSTM algorithm can be applied to improve the long-term navigation capability of the current method.


\bibliographystyle{IEEEtran}
\bibliography{IEEEexample}

\end{document}